%% file: cvpr.tex
\documentclass[final]{cvpr}

\usepackage{times}
\usepackage{epsfig}
\usepackage{graphicx}
\usepackage{amsmath}
\usepackage{amssymb}
\usepackage{subcaption}
\usepackage{multirow}
\usepackage{pifont}
\usepackage{floatrow}
\usepackage{tabularx}
\usepackage{bbm}
\usepackage{float}
\floatstyle{plaintop}
\restylefloat{table}
\usepackage[tableposition=top]{caption}

\newcommand{\cmark}{\ding{51}}%
\newcommand{\xmark}{\ding{55}}%

\usepackage[pagebackref=true,breaklinks=true,colorlinks,bookmarks=false]{hyperref}

\def\cvprPaperID{5} 
\def\confYear{CVPR 2021}

\begin{document}

\title{Video Class Agnostic Segmentation Benchmark for Autonomous Driving}

\author{Mennatullah Siam\\
University of Alberta\\
Edmonton, Canada\\
{\tt\small mennatul@ualberta.ca}
\and
Alex Kendall\\
Wayve\\
London, UK\\
{\tt\small alex@wayve.ai}
\and 

Martin Jagersand\\
University of Alberta\\
Edmonton, Canada\\
{\tt\small jag@cs.ualberta.ca}
}

\maketitle

\begin{abstract}
 \input{core/abstract}
\end{abstract}

\section{Introduction}
\input{core/intro}

\section{Related Work}
\input{core/related}
\input{core/method}

\section{Experimental Results}
\input{core/exps}

\section{Conclusion}
\input{core/conc}

{\small
\bibliographystyle{ieee_fullname}
\bibliography{egbib}
}

\end{document}

%% file: core/abstract.tex
Semantic segmentation approaches are typically trained on large-scale data with a closed finite set of known classes without considering unknown objects. In certain safety-critical robotics applications, especially autonomous driving, it is important to segment all objects, including those unknown at training time. We formalize the task of video class agnostic segmentation from monocular video sequences in autonomous driving to account for unknown objects. Video class agnostic segmentation can be formulated as an open-set or a motion segmentation problem. We discuss both formulations and provide datasets and benchmark different baseline approaches for both tracks. In the motion-segmentation track we benchmark real-time joint panoptic and motion instance segmentation, and evaluate the effect of ego-flow suppression. In the open-set segmentation track we evaluate baseline methods that combine appearance, and geometry to learn prototypes per semantic class. We then compare it to a model that uses an auxiliary contrastive loss to improve the discrimination between known and unknown objects. Datasets and models are publicly released at \url{https://msiam.github.io/vca/}.

%% file: core/intro.tex
\begin{figure}
    \centering
    \includegraphics[width=\textwidth]{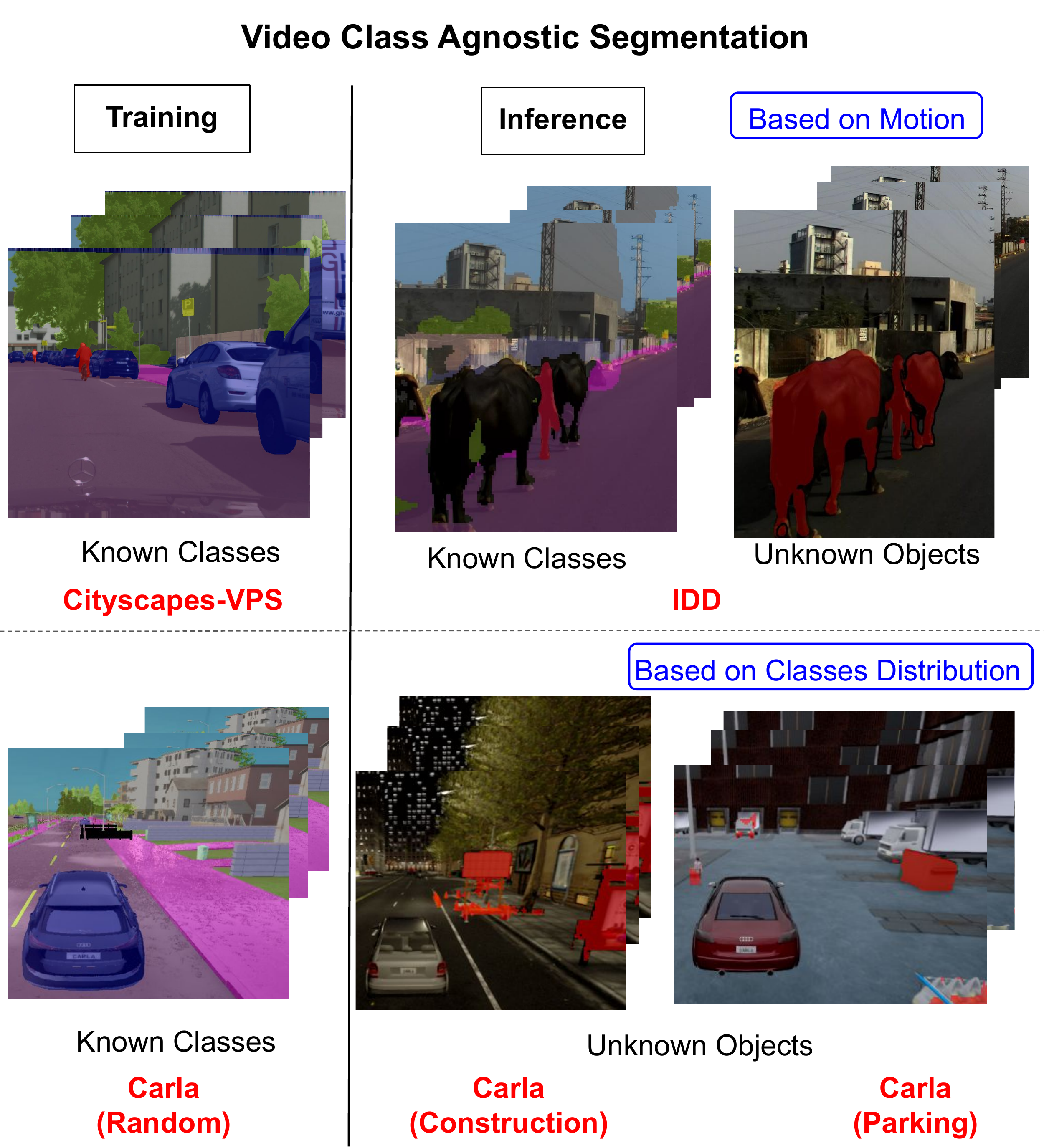}
    \caption{Video Class Agnostic Segmentation enables identification of objects outside the closed set of known classes. Predictions of known and unknown objects on Cityscapes-VPS~\cite{kim2020video}, IDD~\cite{varma2019idd}, and our Synthetic Dataset.}
    \label{fig:teaser}
\end{figure}

Semantic scene understanding is crucial in autonomous driving in both end-to-end and mediated perception approaches as described in~\cite{chen2015deepdriving}. Semantic segmentation which performs pixel-wise classification of the scene is mostly trained on large scale data with closed set of known classes~\cite{cordts2016cityscapes}. A system trained on a limited set of classes would face difficulties in unexpected situations that could occur in different autonomous driving scenarios, such as the animal or parking scenario objects shown in Figure~\ref{fig:teaser}. These scenarios demonstrate potential objects outside the closed set of known classes that are mostly available in public datasets~\cite{cordts2016cityscapes}. Some work was conducted on zero-shot/few-shot learning~\cite{bucher2019zero,wang2019panet} to enable deep neural networks to generalize to novel classes with limited data or side information, but does not address the main problem. For safety reasons, autonomous driving systems should be able to detect objects that are not known beforehand, which are called \textit{``unknown unknown classes''}~\cite{geng2020recent}. 

In this paper we formulate the task of video class agnostic segmentation to account for unknown objects and publicly release the necessary datasets and baselines for the different task formulation. Video class agnostic segmentation is defined as the task of segmenting objects without regards to their semantics, but combining appearance, motion and geometry from monocular video sequences. It is crucial first to identify how to formulate this problem. Since the main goal is to segment obstacles without regards to their semantics, we choose two main tracks for our benchmark: (1) The \textbf{motion segmentation track} that will identify moving objects regardless of their semantics. (2) The \textbf{open-set segmentation track} that will identify pixels belonging to objects outside the closed set of known classes without identifying its exact semantic class. The open-set segmentation formulation is harder than motion segmentation but has the clear advantage of identifying static objects. Examples of both tracks are shown in Figure~\ref{fig:teaser}.

Motion segmentation has been recently studied for autonomous driving in~\cite{siam2017modnet,Rashed_2019_ICCV_Workshops,mohamed2020instancemotseg,Valada_2017_IROS}. Unlike previous literature, we focus on the joint panoptic and motion instance segmentation as a means to improve hollistic scene understanding. We provide motion segmentation labels on both Cityscapes-VPS~\cite{kim2020video} and KITTIMOTS~\cite{Voigtlaender2019CVPR} datasets. Our extended motion segmentation dataset increases the number of labelled sequences and object categories while providing manually labelled instance segmentation masks, unlike previous work that used weakly annotated masks~\cite{mohamed2020instancemotseg}. On the other hand, the open-set segmentation task for autonomous driving has not been thoroughly studied in autonomous driving~\cite{wong2020identifying}. In our work we explore contrastive learning to improve the model's ability to segment known and unknown objects with any baseline model. We propose a contrastive learning method similar to~\cite{winkens2020contrastive}, but their work focused solely on the classification task we extend it to the segmentation of unknown objects. To summarize, our main contributions are:

\begin{itemize}
    \item We formalize the task of \textit{Video Class Agnostic Segmentation} and benchmark different baseline approaches for both motion and open-set segmentation tracks. 
    \item Our motion segmentation dataset offers more labelled sequences and object categories on both KITTI~\cite{Voigtlaender2019CVPR} and Cityscapes~\cite{kim2020video} while providing manually labelled dense masks.
    \item For the open-set segmentation track, we release a large-scale synthetic dataset for different autonomous driving scenarios. These scenarios can be included as part of the Carla Autonomous Driving Challenge~\cite{dosovitskiy2017carla} to drive forward research in both perception and policy learning.
\end{itemize}

%% file: core/related.tex
\textbf{Datasets:} Since our goal is performing \textit{video class agnostic segmentation}, a related task is generic video object segmentation that is not focused on autonomous driving. The Densely Annotated VIdeo Segmentation (DAVIS)~\cite{Perazzi2016,Pont-Tuset_arXiv_2017} dataset provided a benchmark for both unsupervised and semi-supervised video object segmentation tracks. The Freiburg Berkley Motion Segmentation (FBMS)~\cite{brox2010object} and SegTrack~\cite{BMVC.24.56} provided a smaller scale video object segmentation dataset. However, all these generic datasets have much simpler scenes to segment than autonomous driving scenes. They mainly contain few visually salient objects in short video sequences that sometimes can be of low resolution. The camera motion is much simpler than what is experienced in autonomous driving which can significantly affect the methods trained for class agnostic segmentation based on motion. 

More datasets in autonomous driving were recently released for the motion segmentation task~\cite{Valada_2017_IROS,siam2017modnet,Rashed_2019_ICCV_Workshops}. However, they focused solely on vehicles which limits the generalization of the class agnostic segmentation to unknown moving objects. We resolve this issue in our provided dataset for motion segmentation. Recently, the Rare4D~\cite{wong2020identifying} dataset was collected for LIDAR data to segment different rare and unknown objects. However, the dataset was not publicly released and is focused on LIDAR data. We release the first class agnostic segmentation dataset focused on monocular video sequences from cameras, and provide further analysis on the relation between objects labelled as unknown between the training and testing phases.

\textbf{Motion Segmentation:} has been recently studied in the literature of autonomous driving with supervised learning methods~\cite{Valada_2017_IROS,siam2017modnet,Rashed_2019_ICCV_Workshops,mohamed2020instancemotseg}. Vertens et. al.\cite{Valada_2017_IROS} proposed a joint model for semantic and motion segmentation with ego-flow suppression, and released Cityscapes-KITTI-Motion dataset for only car category. Siam et. al.\cite{siam2017modnet} concurrently released one of the earliest motion segmentation dataset KITTIMoSeg, that was extended by Rashed et al.~\cite{Rashed_2019_ICCV_Workshops} and with instance labels by Mohammed et. al.~\cite{mohamed2020instancemotseg}. In our work we mainly focus on extending motion segmentation datasets to include more object categories to have eight object categories instead of only \textit{car} and increase the number of sequences by 50$\times$. These help to improve the generalization ability required to enable the segmentation of unknown moving objects. Our dataset also provides motion annotations on both KITTI and Cityscapes instead of focusing only on KITTI as~\cite{mohamed2020instancemotseg}.  We also improve the annotations by providing manually labelled segmentation masks, unlike weakly annotated ones~\cite{mohamed2020instancemotseg}. Additionally, the corresponding tracking labels and panoptic segmentation labels from the original datasets are provided. A full comparison of these datasets is shown in Table~\ref{table:datasets_motion}.

\textbf{Unknown Objects Segmentation:} in autonomous driving has not been thoroughly studied in the literature with only two recent works~\cite{wong2020identifying,ovsep20204d}. Wong et. al.~\cite{wong2020identifying} proposed a method inspired by~\cite{snell2017prototypical} and trained on TOR4D, which is a large scale LIDAR dataset, and presented the Rare4D dataset to evaluate unknown objects. However, Rare4D dataset is not publicly available and shares no indication of the relation between unknown objects labelled between the training and testing datasets. Osep et al.~\cite{ovsep20204d} proposed a method that uses a video sequence of stereo images to predict 4D generic proposals for autonomous driving. But they evaluate only on 150 images for the autonomous driving dataset with unknown objects, which is insufficient to evaluate model scalability. They build their model on a two-stage instance segmentation method, which is computationally inefficient, while most of the unknown objects belong to stuff classes such as cones, traffic warning, and barriers. These classes do not require the separation of instances as they will not be tracked. These limitations in the literature drove us to construct a benchmark for monocular video sequences for the open-set unknown objects track.

%% file: core/method.tex
\section{Motion Segmentation Track}
\subsection{Dataset}

\begin{table*}[t]
\caption{Comparison of different datasets for motion segmentation. M: stands for manually labelled annotations, W: stands for weakly labelled annotations. Seqs: sequences, Cats: categories. Our dataset provides instance-wise manually labelled annotations for moving objects and increases the sequences and categories unlike previous datasets.}
\label{table:datasets_motion}
\centering
\begin{tabular}{|l|l|l|l|l|l|l|l|}
\hline
Dataset & \# Frames & \# Seqs & \# Cats & Instances & Panoptic & Tracks & Annotation\\ \hline
 DAVIS~\cite{caelles20192019} & 6208 & 90 & 78 & \cmark & \xmark & \cmark & M\\ \hline
 KITTI Motion~\cite{Valada_2017_IROS} & 455 & - & 1 (Car Only) & \xmark & \xmark & \xmark & M\\
 KITTI-MoSeg~\cite{siam2017modnet}\cite{Rashed_2019_ICCV_Workshops} & 12919 & $\sim$ 38 & 1 (Car Only) & \xmark & \xmark & \xmark & W\\
 InstKITTI-MoSeg\cite{mohamed2020instancemotseg} & 12919 & $\sim$ 38 & 5 & \cmark & \xmark & \xmark & W\\
 Cityscapes Motion\cite{Valada_2017_IROS} & 3475 & - & 1 (Car Only) & \cmark & \cmark & \xmark & M\\ \hline
 VCAS-Motion (ours) & 11008 & \textbf{520} & \textbf{8} & \cmark & \cmark & \cmark & \textbf{M}\\ \hline
\end{tabular}
\end{table*}

\begin{figure*}[t]
\centering
\begin{subfigure}{.5\textwidth}
    \includegraphics[width=\textwidth]{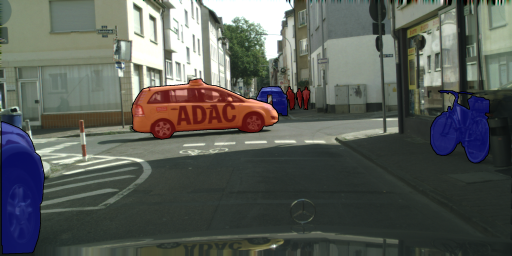}
\end{subfigure}%
\begin{subfigure}{.5\textwidth}
    \includegraphics[width=\textwidth]{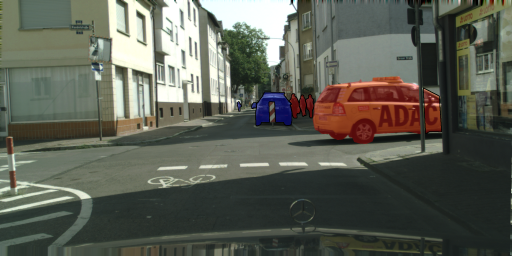}
\end{subfigure}

\begin{subfigure}{.5\textwidth}
    \includegraphics[width=\textwidth]{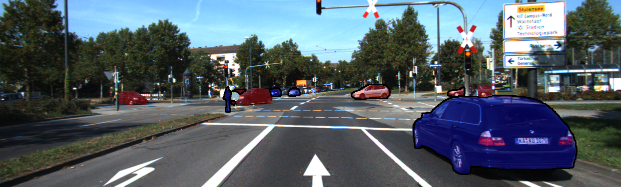}
\end{subfigure}%
\begin{subfigure}{.5\textwidth}
    \includegraphics[width=\textwidth]{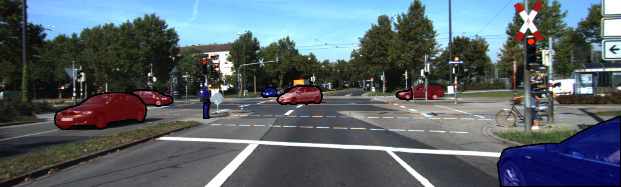}
\end{subfigure}
\caption{Our Extended Motion Annotations showing two consecutive frames on: (a) Top Row: Cityscapes datasets and (b) Bottom Row: KITTI. Red: moving instances, Blue: static instances.}
\label{fig:wayve_Dataset}
\end{figure*}

For motion segmentation we provide motion annotations to extend the original KITTIMOTS~\cite{Voigtlaender2019CVPR} and Cityscapes-VPS~\cite{kim2020video} datasets as shown in Figure~\ref{fig:wayve_Dataset}. A trajectory annotation tool was built to annotate the trajectories for either moving or static and is further used to provide instance-wise motion masks. Table~\ref{table:datasets_motion} shows the different statistics for our dataset in comparison to others. Our dataset has the main advantage of providing a larger variety of object categories and 50$\times$ increase in video sequences. These substantially improve the scalability of video class agnostic segmentation. Crucially, this pushes models to depend more on motion information and only use appearance information to detect objectness rather than learning to semantically detect cars only.

Although DAVIS provides a larger variety of object categories, the camera motion and scenes in these sequences are much simpler and easier to segment unlike our autonomous driving setting. We also provide instance masks unlike most previous literature which is crucial for motion instance segmentation. Concurrent to our work~\cite{mohamed2020instancemotseg} provided instance-wise motion masks, however we provide manually labelled segmentation masks unlike their weakly annotated ones. More importantly we provide annotations for both KITTI and Cityscapes, while their dataset is built solely on KITTI. This can affect the generalization ability of the class agnostic segmentation. Our dataset also benefits from the already provided panoptic labels for 3000 frames and tracking annotations for all frames which can aid in tracking moving objects. The statistics of moving and static instances for both datasets are shown in Table~\ref{table:stats}.

\begin{figure*}[t]
\begin{minipage}{0.32\linewidth}
\centering
\begin{tabular}{|l|l|}
\hline
             Scenario     & Unknown Objects \\ \hline
\multirow{3}{*}{Parking} & Cart with Bags \\
                  &  Shopping Trolley\\
                  &  Garbage Bin\\ \hline
\multirow{3}{*}{Construction} &  Traffic Warning\\
                  &  Construction Cone\\ \hline 
Barrier & Traffic Pole\\ \hline
\multirow{5}{*}{Training} &  Barrel\\
                          & Traffic Cone \\ 
                          & Traffic Barrier\\ 
                          & Static (others) \\
                          & Dynamic (others) \\ \hline
\end{tabular}%
\end{minipage}%
\begin{minipage}{0.68\linewidth}
\centering
\centering
\begin{subfigure}{.28\textwidth}
    \includegraphics[width=\textwidth]{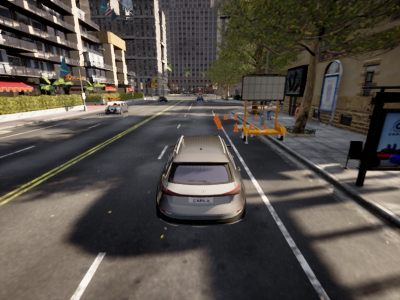}
\end{subfigure}%
\begin{subfigure}{.28\textwidth}
    \includegraphics[width=\textwidth]{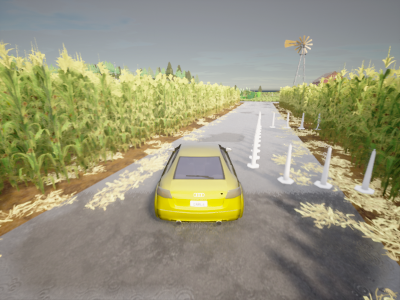}
\end{subfigure}%
\begin{subfigure}{.28\textwidth}
    \includegraphics[width=\textwidth]{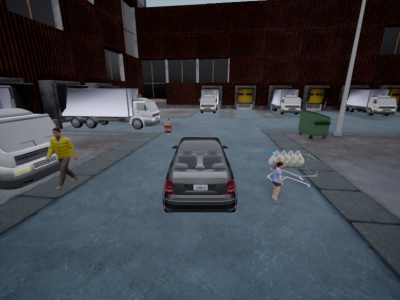}
\end{subfigure}

\begin{subfigure}{.28\textwidth}
    \includegraphics[width=\textwidth]{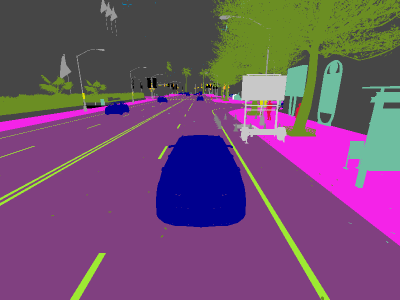}
\end{subfigure}%
\begin{subfigure}{.28\textwidth}
    \includegraphics[width=\textwidth]{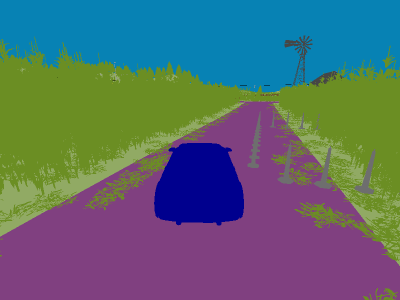}
\end{subfigure}%
\begin{subfigure}{.28\textwidth}
    \includegraphics[width=\textwidth]{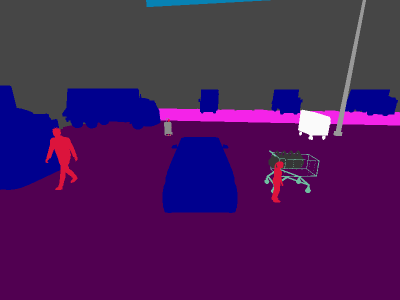}
\end{subfigure}
\end{minipage}
\caption{Different scenarios in Carla Simulation and objects considered as unknown in our synthetic data.}
\label{table:carla}
\end{figure*}

\begin{table}[h]
\centering
\caption{Dataset Statistics of moving and static instances.}
\label{table:stats}
\begin{tabular}{|c|c|c|}
\hline
& KITTI & Cityscapes \\ \hline
Moving & 5788 & 41334\\
Static & 6094 & 43981\\\hline
\end{tabular}
\end{table}

\subsection{Baselines and Metrics}
We propose a real-time two-stream multitask model that performs panoptic and motion instance segmentation as a baseline model. Our backbone network is a two-stream ResNet-50~\cite{he2016deep} with a one-way feature pyramid network~\cite{lin2017feature}. We extend the real-time instance segmentation model from SOLO~\cite{wang2019solo} with a semantic segmentation head to segment the \textit{stuff} classes and a class agnostic head that segments moving instances. The class agnostic head provides redundant signal for a safety-critical approach. Our panoptic segmentation model is initially trained on Cityscapes~\cite{cordts2016cityscapes} and Cityscapes-VPS~\cite{kim2020video} datasets. Then the class agnostic head is trained on our motion segmentation datasets with fixed weights for the rest of the model. The input to the two-stream model is both appearance and motion as optical flow encoded in RGB. We compare two baselines the one with input optical flow estimated directly from Flownet-2~\cite{ilg2017flownet}, while the other takes ego-flow suppressed one. The ego flow suppression uses the estimated depth and pose from~\cite{godard2019digging}, then computes the ego flow following the method from~\cite{Valada_2017_IROS}. The ego-flow suppressed output is the result of subtracting the ego flow from the original estimated flow. The instance segmentation and class agnostic segmentation heads are trained with focal loss~\cite{lin2017focal}, and dice loss~\cite{milletari2016v}, while the segmentation head is trained with cross entropy. 

\begin{figure*}[t]
\centering
    \includegraphics[width=0.9\textwidth]{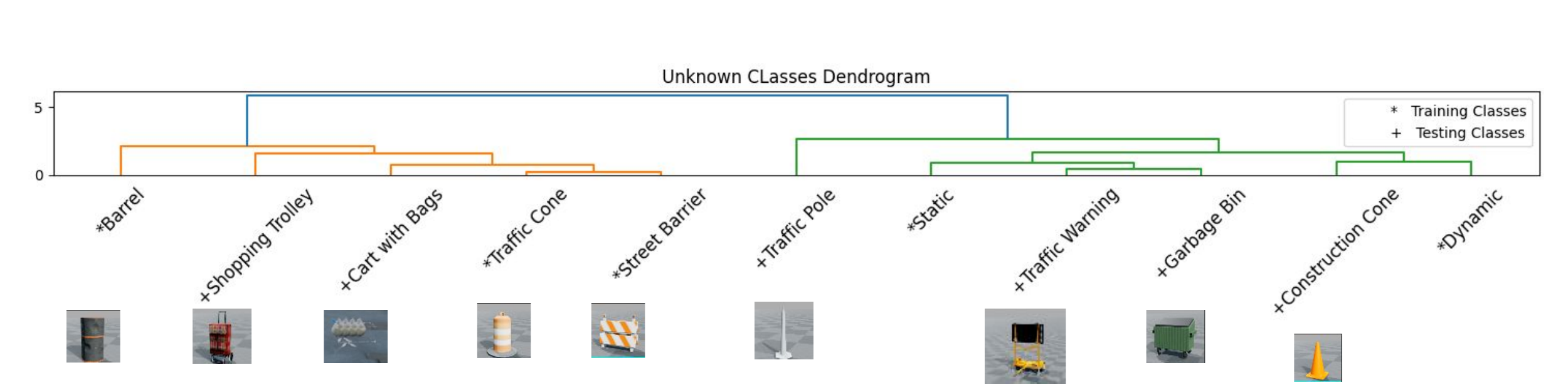}
    \caption{Dendrogram among unknown objects used during training and testing phases.}
    \label{fig:dendrogram}
\end{figure*}

We use the same evaluation metrics provided by~\cite{wong2020identifying} for measuring unknown instance segmentation. We report the class agnostic quality metric as shown in~\ref{eq:caq}, which combines the segmentation and recognition quality. $\text{TP}$ and $\text{FN}$ are the true positives and false negatives respectively. Only instance masks that have intersection over union with the ground-truth above 0.5 are considered true positives.

\begin{subequations}
\begin{equation}
    \text{SQ} = \frac{\sum\limits_{p, g \in \text{TP}} \text{IoU}(p, g)}{|\text{TP}|}
\end{equation}
\begin{equation}
    \text{RQ} = \frac{|\text{TP}|}{|\text{TP}| + |\text{FN}|}
\end{equation}
\begin{equation}
    \text{CAQ} = \text{SQ}.\text{RQ}
\end{equation}
    \label{eq:caq}
\end{subequations}

\section{Open-set Segmentation Track}
In this section we discuss the synthetic dataset that we curated to conduct controlled experiments with a large-scale dataset to ensure the generalization ability of class agnostic segmentation of unknown objects.

\subsection{Carla Scenarios and Dataset}
In the open-set segmentation formulation we care about providing video sequences along with annotations for unknown objects in different autonomous driving scenarios. Thus, we build different scenarios within the Carla simulation environment~\cite{dosovitskiy2017carla}. The goal is to incorporate these scenarios as part of the Carla challenge for autonomous driving~\footnote{See the Carla challenge here: \url{https://leaderboard.carla.org/challenge/}} to benefit both perception and policy learning researchers. Such scenarios serve as a way to evaluate the robustness of autonomous driving systems and the safety-critical design. Additionally, the available video dataset with dense labels (Cityscapes-VPS) has 3000 images and 300 for evaluation. Thus we build our aggregate data with approximately 70K frames. 

\begin{figure*}[t!]
\centering
\includegraphics[width=0.78\textwidth]{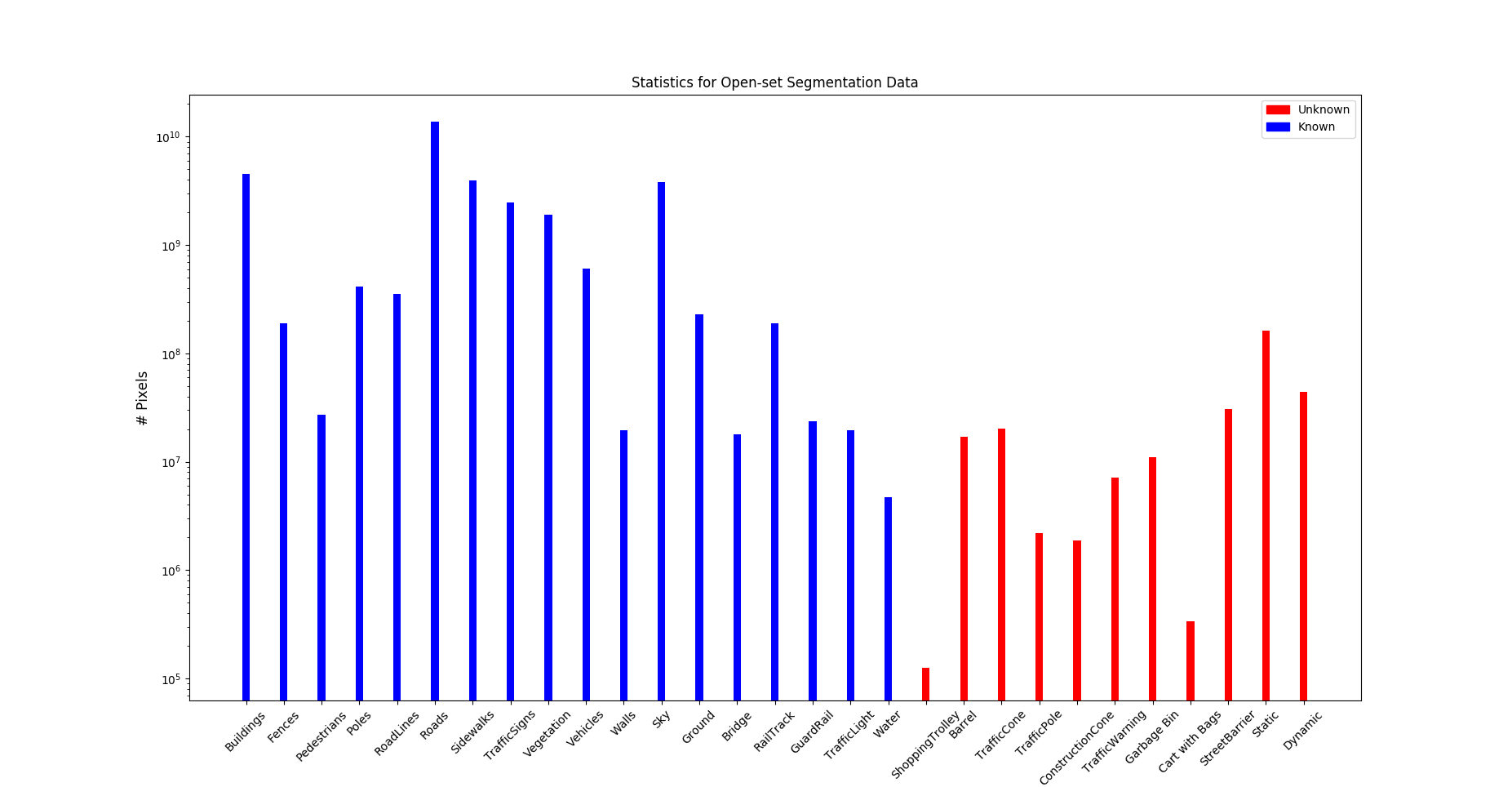}
\caption{Statistics for our Carla data for both known classes and unknown objects showing number of pixels per class.}
\label{fig:carla_stats}
\end{figure*}

First, we insert unknown objects and modify the basic virtual driving agent in Carla to avoid obstacles, in order to collect large-scale data with ground-truth depth and semantic segmentation labels. We then randomize object placement, weather condition, traffic (vehicles and pedestrian) and use different unknown objects and towns between training and testing as shown in Table~\ref{table:openset}. Figure~\ref{table:carla} lists the three main scenarios that are used to evaluate the open-set segmentation task, and Figure~\ref{fig:carla_stats} shows the class statistics.

\begin{table}[h]
\caption{Towns and Images collected in both training and testing splits for our open-set track in Carla.}
\label{table:openset}
\centering
\begin{tabular}{|c|c|c|}
\hline
& Training & Testing \\ \hline
\multirow{3}{*}{Towns} & Town1 &  Town10\\ 
 & Town2 & Town7 \\ 
 & Town3 & \\\hline
\# Images & 57972 & 1886\\\hline
\end{tabular}
\end{table}

\subsection{Analysis on the Unknown Objects Relations}
\label{sec:unknown}
We extend the carla environment to provide fine-grained labels for the specific set of unknown objects that we placed. The fine-grained labels are used to analyze the relation among unknown objects used during training and testing. This provides better understanding of task difficulty and model scalability. We propose to use the prototype features from masked average pooling following equation~\ref{eq:map}, where $X_i$ is the $i^{th}$ input image, $F_i$ is the corresponding feature maps from ResNet-50~\cite{he2016deep} with pretrained ImageNet weights $\theta$, and $M_i$ is its segmentation mask. $P_c$ denotes the prototype for class $c$, that is computed on $N$ images within the dataset. A pair-wise distance $d(i, j)$ is measured among classes followed by agglomerative clustering.

A dendrogram among the classes is visualised to understand the relation between the unknown objects in the training and testing phases as shown in Figure~\ref{fig:dendrogram}. Classes that are visually similar, especially in texture, are closer such as street barrier and traffic cone. From the figure we can see less correlation among some of the different unknown objects used during testing and training especially in the parking scenario. Later, in the experiments we evaluate per scenario to confirm the generalization ability of the model.

\begin{subequations}
\begin{equation}
    F_i = f_{\theta}(X_i)
\end{equation}
\begin{equation}
    P_c = \sum\limits_{i=1}^N \sum\limits_{x, y} \mathbbm{1}[M_i^{x, y}=c] F_i^{x, y}
\end{equation}
\begin{equation}
    d(i, j) = \left\lVert P_i - P_j \right\rVert_2
\end{equation}
\label{eq:map}
\end{subequations}


\subsection{Baselines and Metrics}
The method we use for the open-set segmentation is similar to~\cite{wong2020identifying} but focused only on semantic segmentation without incorporating instances. We learn a representative prototype $\mu, \sigma$ per class. Then we use the distance to classify every pixel based on the matching prototype in~\ref{eq:dist}. We use a baseline that relies on both appearance and geometry, as we found depth to be an important signal to identify unknown objects. A two-stream backbone model $f_{\theta}$ with ResNet-50 backbone~\cite{he2016deep} and a feature pyramid network~\cite{lin2017feature}, which takes as input appearance and depth. This is followed by a semantic segmentation head $f_{\phi}$ with 4 convolutional modules with ReLU and group normalization, that learns prototypes $\mu, \sigma$ per class.  

Let the input appearance and depth be denoted as $x$, the extracted features $h = f_{\theta}(x)$, and the embeddings from the segmentation head $m=f_{\phi}(h)$. The semantic segmentation head predicts the class probabilities following equation~\ref{eq:mahala}. Where the distance $d_{i, k}$ denotes the distance of pixel $i$ features to the representative prototype of class $k$. A global learnable constant $\gamma$ is used to estimate the unknown objects regardless of the objects' semantics similar to~\cite{wong2020identifying}, then a softmax over $C+1$ distances is used to estimate the probability of the pixel to belong to a certain class. For the contrastive learning baseline we learn a separate projection head with auxiliary contrastive loss along with the segmentation loss as shown in equation~\ref{eq:loss}.

\begin{subequations}
    \begin{equation}
        d_{i, k} = \frac{- \left\lVert m_i - \mu_k \right\rVert^2}{2 \sigma_k^2} 
        \label{eq:dist}
    \end{equation}
    \begin{equation}
        d_{i, C+1} = \gamma
    \end{equation}
    \begin{equation}
        \hat{y}_{i, k} = \frac{\exp(d_{i, k})}{\sum\limits_{j=1}^{C+1}{\exp(d_{i, j})}}
    \end{equation}
    \begin{equation}
        l_{seg} = \frac{-1}{N} \sum\limits_{i=1}^N \sum\limits_{k=1}^{C+1}y_{i, k} \log{\hat{y}_{i, k}}
    \end{equation}
    \label{eq:mahala}
\end{subequations}

\begin{equation}
    L = L_{seg} + \lambda L_{cl}
    \label{eq:loss}
\end{equation}

As for the evaluation metric we report intersection over union in a class agnostic manner over the unknown objects in the test set. Where the predicted mask using the global learnable constant $\gamma$ is used to label objects outside the closed set of known classes.

%% file: core/exps.tex
\subsection{Experimental Setup}
\begin{table*}[t]
\centering
\caption{Results of class agnostic and panoptic segmentation model. Tr-Te: Training and Testing motion segmentation datasets used. EFS: Ego Flow Suppression. D: DAVIS, K: KITTI our annotations, C: Cityscapes our annotations.}
\label{table:vcanet-motion}
\begin{tabular}{|l|c|c|c|c|c|c|c|c|}
\hline
 Model & Tr-Te & EFS & \multicolumn{3}{c|}{CA Metrics} & \multicolumn{3}{c|}{Panoptic Metrics} \\ \hline
 &  &  & SQ & RQ & CAQ  & $PQ_{All}$  & $PQ_{Th}$ & $PQ_{St}$ \\ \hline
 Pan-Base & - & - & - & - & - & 56.4  & 49.9  & 61.1 \\
 CA-Base  & D-D & \xmark & 76.6 & 60.6 &  46.4 & -  & -  & -\\
 VCANet & D-D & \xmark & 72.2 & 53.2 & 38.6 & 56.4  & 49.9  & 61.1\\ \hline
 VCANet & KC-K & \xmark & 82.4 & 72.2 & 59.5 & 56.4  & 49.9  & 61.1\\
 VCANet & KC-K & \cmark & \textbf{82.4} & \textbf{75.4} & \textbf{62.1} & 56.4  & 49.9  & 61.1\\ \hline
 VCANet & KC-C & \xmark & 77.9 & \textbf{57.3} & \textbf{44.7} & 56.4  & 49.9  & 61.1\\
 VCANet & KC-C & \cmark & \textbf{78.0} & 56.2 & 43.8 & 56.4  & 49.9  & 61.1\\ \hline
\end{tabular}
\end{table*}

\begin{figure*}[t]
\centering
\begin{subfigure}{.45\textwidth}
    \includegraphics[width=\textwidth]{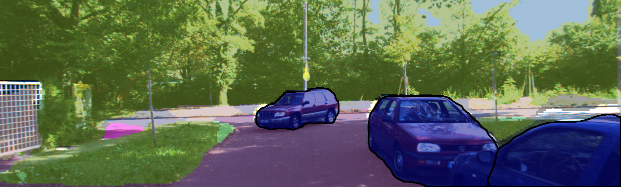}
\end{subfigure}%
\begin{subfigure}{.26\textwidth}
    \includegraphics[width=1.05\textwidth]{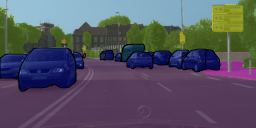}
\end{subfigure}%
\begin{subfigure}{.28\textwidth}
    \includegraphics[width=0.865\textwidth]{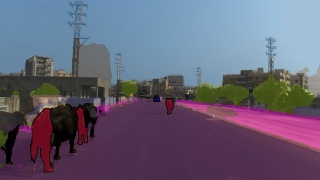}
\end{subfigure}

\begin{subfigure}{.45\textwidth}
    \includegraphics[width=\textwidth]{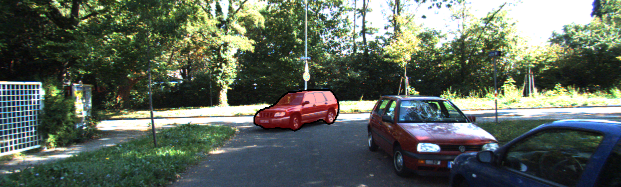}
    \caption{}
\end{subfigure}%
\begin{subfigure}{.26\textwidth}
    \includegraphics[width=1.05\textwidth]{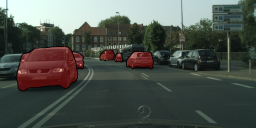}
    \caption{}
\end{subfigure}%
\begin{subfigure}{.28\textwidth}
    \includegraphics[width=0.865\textwidth]{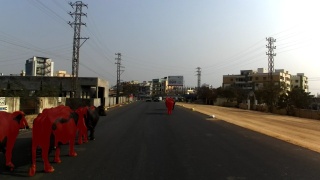}
    \caption{}
\end{subfigure}
\caption{Predicted panoptic and CA segmentation on (a) KITTI. (b) Cityscapes. (c) IDD. Top: panoptic segmentation. Bottom: class agnostic segmentation.}
\label{fig:motionpred}
\end{figure*}

\textbf{Open-set Segmentation Setup:}
Throughout all experiments we use SGD with momentum optimizer with 0.005 learning rate and 0.9 momentum, and weight decay of $1 \times 10^{-4}$ for 50 or 15 epochs for Cityscapes-VPS or Carla datasets respectively. A step learning rate scheduling which reduces the learning rate by 0.1 at epochs 30, 40 for Cityscapes-VPS, and 6, 8 for Carla is used. On Cityscapes-VPS we resize images to $1024 \times 512$ then use random augmentations as random scales $\{0.8, 1.3\}$, random flipping and random cropping with $320 \times 512$ as crop sizes.

For datasets, we mainly use Cityscapes-VPS~\cite{kim2020video} as it has dense annotations for the video sequences and our collected dataset on Carla. In order to ensure no overfitting for the unknown objects segmentation occurs, we use different set of objects labelled as unknown during training and testing phases. In all experiments we include depth as another input modality to better segment unknown objects, in Carla we use the simulation's groundtruth and for Cityscapes-VPS we use the estimated depth from~\cite{godard2019digging}.

\textbf{Motion Segmentation Setup:}
We use similar optimizer and learning rate to the open-set segmentation track, but train for 15 epochs on the combined KITTI-MOTS and Cityscapes-VPS motion datasets. Both datasets have their own training and testing splits, we train on the combined training splits then evaluate on the testing split from each. A step learning rate scheduling which reduces the learning rate with 0.1 at epochs 6, 8 is used. We report panoptic quality~\cite{kirillov2019panoptic} and the class agnostic quality~\cite{wong2020identifying}.

\begin{figure*}[t]
\begin{minipage}{0.2\linewidth}
\centering
\begin{tabular}{|l|c|}
\hline
  Scenario   & CA-IoU  \\ \hline
 Barrier & 22.1\\
 Construction & 41.4 \\
 Parking & 27.0\\\hline
\end{tabular}
\end{minipage}%
\begin{minipage}{0.75\linewidth}
\centering
\begin{subfigure}{.3\textwidth}
    \includegraphics[width=\textwidth]{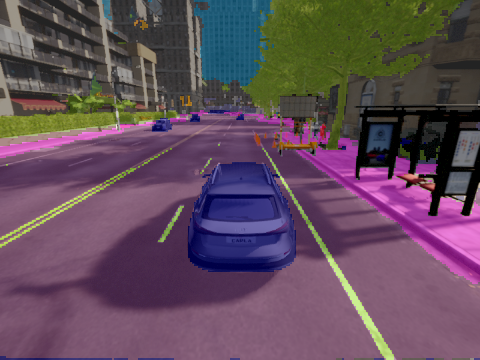}
\end{subfigure}%
\begin{subfigure}{.3\textwidth}
    \includegraphics[width=\textwidth]{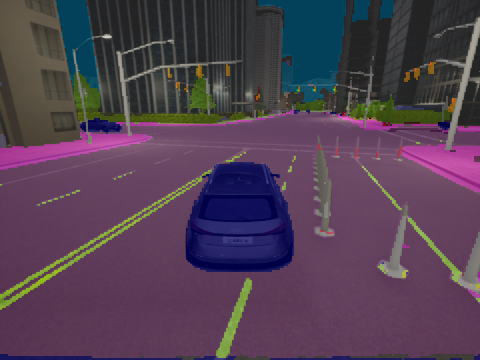}
\end{subfigure}%
\begin{subfigure}{.3\textwidth}
    \includegraphics[width=\textwidth]{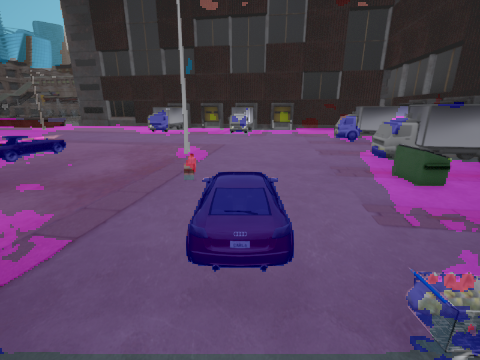}
\end{subfigure}

\begin{subfigure}{.3\textwidth}
    \includegraphics[width=\textwidth]{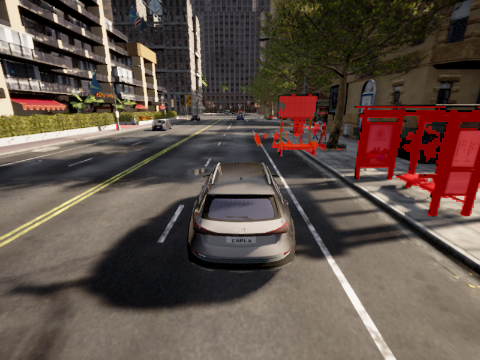}
    \caption{}
\end{subfigure}%
\begin{subfigure}{.3\textwidth}
    \includegraphics[width=\textwidth]{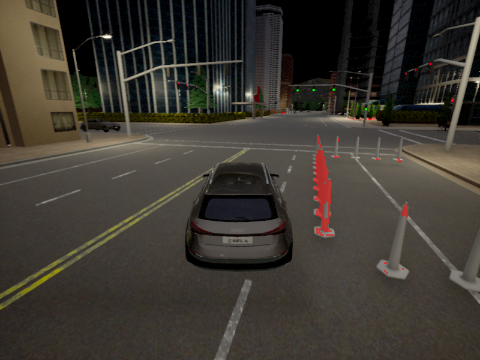}
    \caption{}
\end{subfigure}%
\begin{subfigure}{.3\textwidth}
    \includegraphics[width=\textwidth]{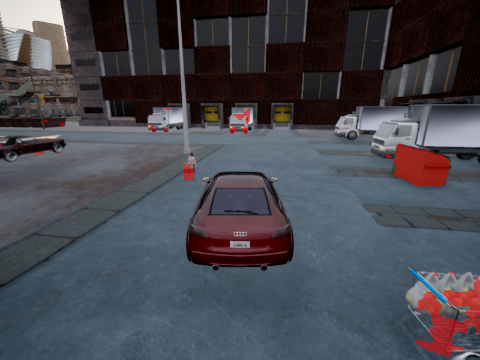}
    \caption{}
\end{subfigure}
\end{minipage}
\caption{CA-IoU reported per scenario. Predicted semantic and class agnostic segmentation on Carla Scenarios (a) Construction. (b) Barrier. (c) Parking. Top: semantic segmentation. Bottom: class agnostic segmentation}
\label{fig:opensetpred_carla}
\end{figure*}

\begin{figure*}[t]
\centering
\begin{subfigure}{.3\textwidth}
    \includegraphics[width=\textwidth]{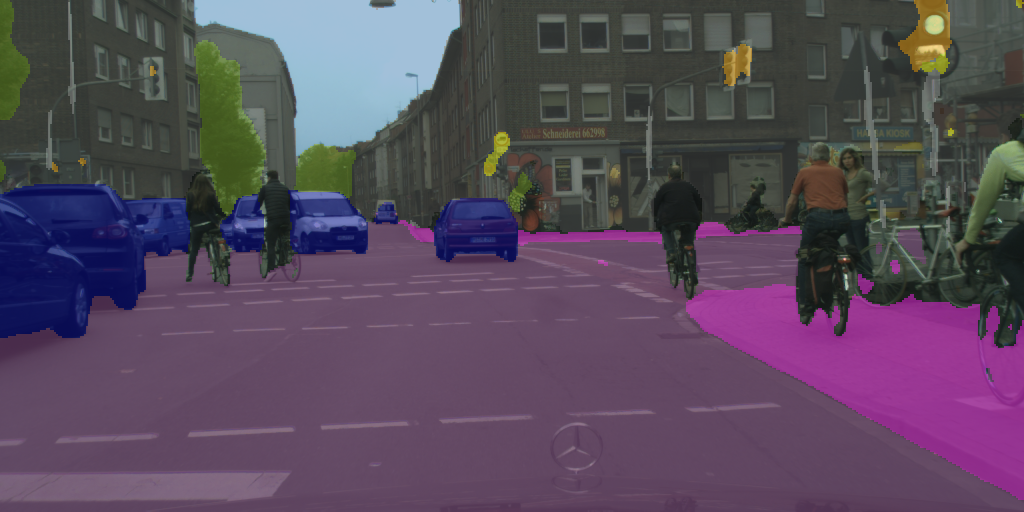}
\end{subfigure}%
\begin{subfigure}{.3\textwidth}
    \includegraphics[width=\textwidth]{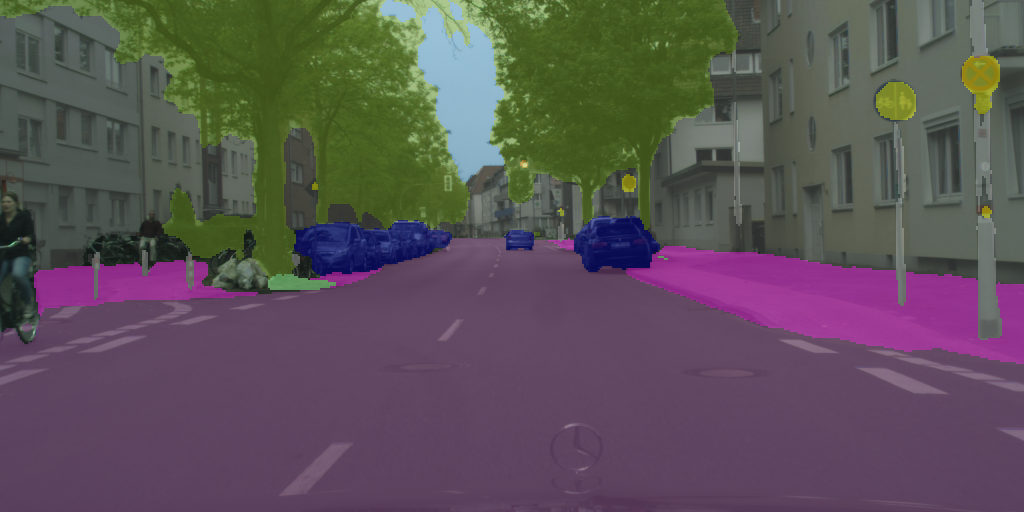}
\end{subfigure}%
\begin{subfigure}{.3\textwidth}
    \includegraphics[width=\textwidth]{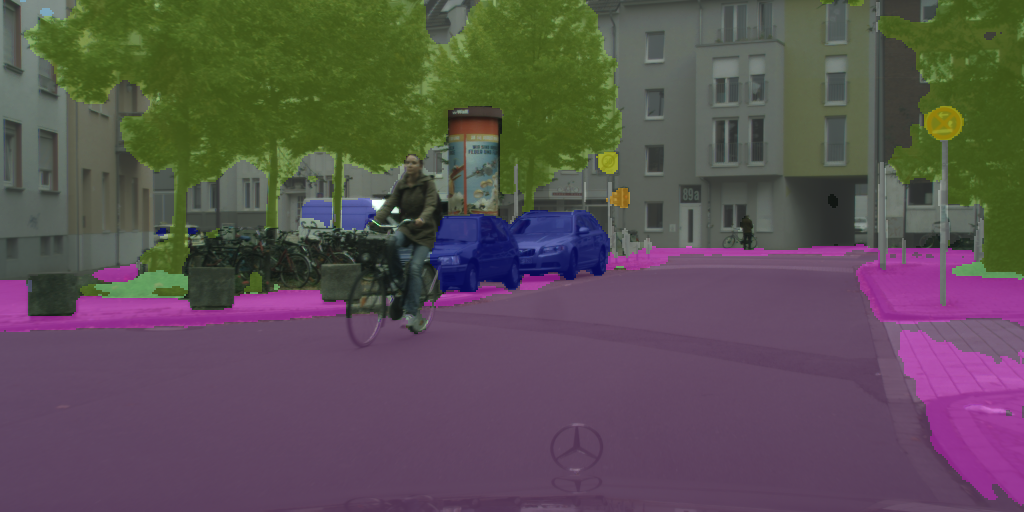}
\end{subfigure}

\begin{subfigure}{.3\textwidth}
    \includegraphics[width=\textwidth]{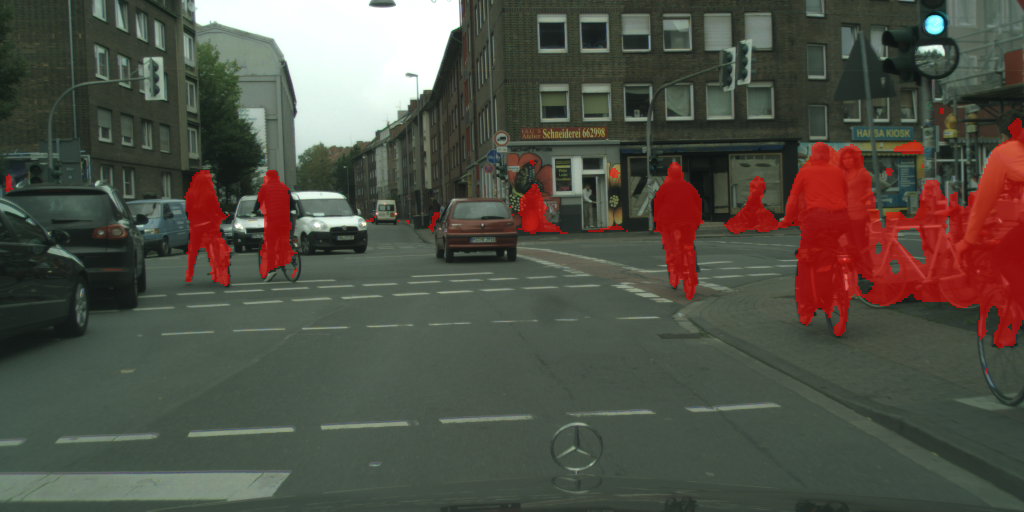}
\end{subfigure}%
\begin{subfigure}{.3\textwidth}
    \includegraphics[width=\textwidth]{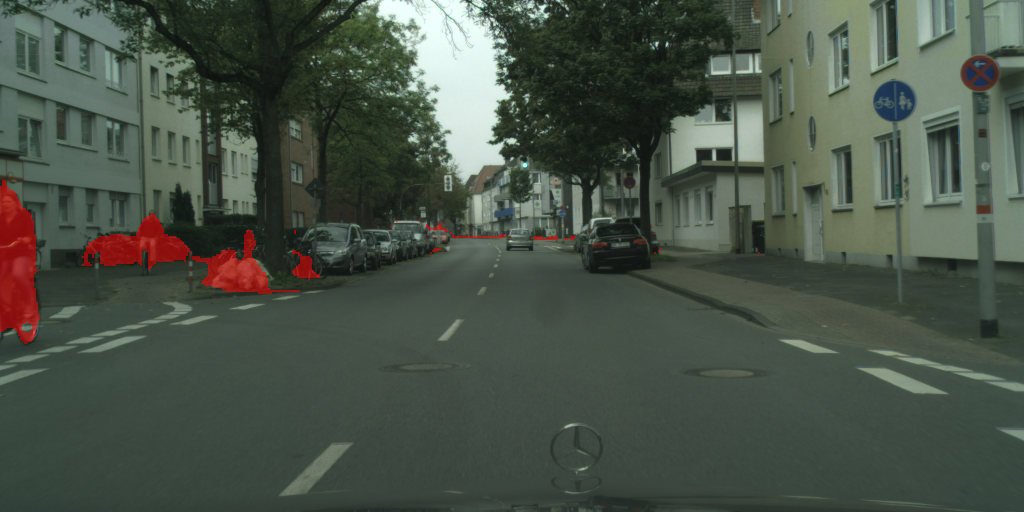}
\end{subfigure}%
\begin{subfigure}{.3\textwidth}
    \includegraphics[width=\textwidth]{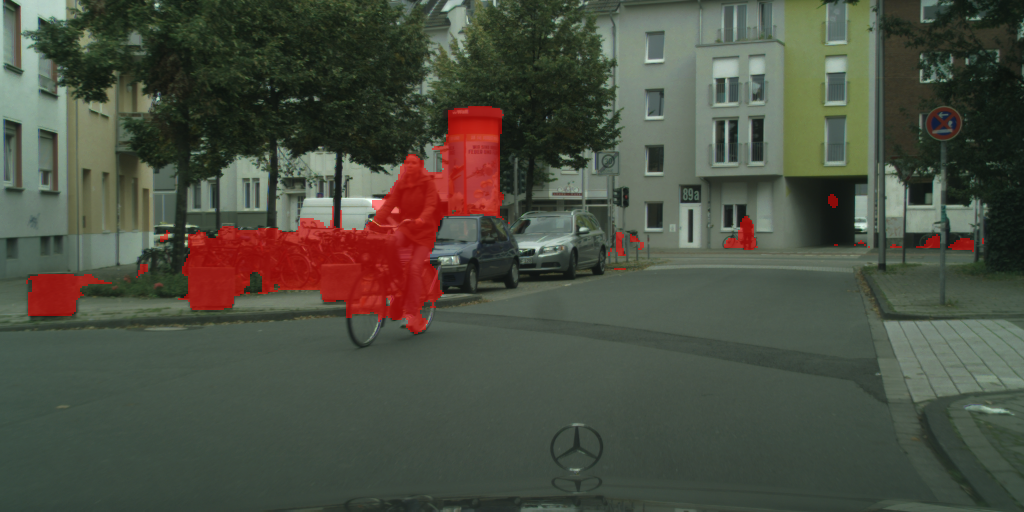}
\end{subfigure}
\caption{Predicted semantic and class agnostic segmentation on Cityscapes-VPS. Top: semantic segmentation. Bottom: class agnostic segmentation (Note: pedestrian - rider - bicycle - motorcycle are left out from training for known classes).}
\label{fig:opensetpred_cscapesvps}
\end{figure*}

\subsection{Motion Segmentation Track Results} 
Table~\ref{table:vcanet-motion} shows the results of our proposed multi-task panoptic and class agnostic segmentation which we term as VCANet. The Pan-Base denotes our panoptic baseline without performing class agnostic segmentation. We initially report the class agnostic baseline trained on DAVIS~\cite{caelles20192019} and compare it with the multi-task model VCANet to show that it maintains comparable segmentation quality. Our VCANet model with ego-flow suppression improves further the class agnostic quality (CAQ) metric on KITTIMOTS motion data. Our results act as a baseline for different approaches that tackle class agnostic instance segmentation. Figure~\ref{fig:motionpred} shows the qualitative analysis of VCANet on CityscapesVPS, KITTIMOTS and IDD. The model trained on our motion data is still able to segment moving objects in IDD sequences, which has an entirely different data distribution, and unknown objects such as animals that never appeared during training on Cityscapes-VPS known classes. Our multi-task model runs real-time at 5 fps on a 1080Ti GPU with $1024 \times 512$ image resolution.

\subsection{Open-Set Segmentation Track Results}
We conduct initial experiments on Cityscapes-VPS where four classes are withheld to be considered as unknown: Person, Rider, Motorcycle, Bicycle. During training, we label Person, Rider, and some of the ignored classes in Cityscapes as unknown objects for training the global unknown constant. Pixels belonging to Motorcycle and Bicycle are ignored in the cross entropy loss and do not contribute to the back-propagation. In the testing phase we evaluate on Motorcycle and Bicycle as the unknown objects and ignore the rest. Table~\ref{table:vca} shows the results for two baseline models that follow the segmentation baseline we presented earlier. We found that using an auxiliary contrastive loss improves the results for the CA-IoU and mIoU metrics on Cityscapes-VPS. Figure~\ref{fig:opensetpred_cscapesvps} shows the results for segmenting both known classes and unknown objects, which demonstrates the ability to segment unknown bicycle and motorcycle classes.

In the same Table~\ref{table:vca} we compare the baseline model with and without contrastive learning and train on our collected synthetic dataset. The CA-IoU is reported only on the unknown objects used during testing, which are labelled in Figure~\ref{fig:dendrogram}. We find that on the large-scale synthetic Carla dataset the baseline with contrastive learning does not improve, which could be explained because Carla dataset size is 25$\times$ of Cityscapes-VPS, and has more pixels labelled as unknown. We conduct experiments on a reduced version of the Carla dataset with only 2400 images as Cityscapes-VPS and less number of pixels labelled as unknown. This becomes more challenging to segment unknown objects and we see a clear gain from the auxiliary contrastive loss in low-data settings. This leaves an open question of how to devise ways to allow contrastive learning to improve performance even with abundant data. Figure~\ref{fig:opensetpred_carla} shows the CA-IoU over different scenarios and their qualitative results. It shows the model is able to segment some of the unknown objects that did not appear during training such as in the Parking Scenario. It is worth noting that some of the objects (Garbage Bin and Traffic Warning) have less correlation with unknown objects used during the training phase as shown in Figure~\ref{fig:dendrogram}. This confirms the model can scale to unknown objects not previously seen during training.

\begin{table}[!t]
\centering
\caption{Open-Set Segmentation Results on Cityscapes-VPS and Carla. CL: contrastive learning. CA-IoU: class agnostic IoU on unknown objects.}
\label{table:vca}
\begin{tabular}{|l|l|l|c|c|c|c|}
\hline
     & Dataset & mIoU & CA-IoU  \\ \hline
 No CL & Cityscapes-VPS & 57.1 & 19.1 \\
 CL &  Cityscapes-VPS & \textbf{62.8} & \textbf{20.9} \\ \hline
 No CL & Carla & \textbf{45.7} & \textbf{41.9} \\
 CL & Carla & 44.2 & 37.2\\\hline
No CL & Carla Reduced & 38.5 & 6.5\\
CL & Carla Reduced & \textbf{41.7} & \textbf{16.0}\\\hline
\end{tabular}
\end{table}

%% file: core/conc.tex
In this paper we formalized the video class agnostic segmentation task and provided necessary datasets and benchmarks for the two main tracks: (1) motion segmentation track, and (2) open-set segmentation track. In the motion segmentation track, an improved dataset for motion instance segmentation is provided and multiple motion segmentation baselines are benchmarked and publicly released. For the open-set segmentation track our synthetic dataset provides the means to assess model scalability to unknown objects and further study the relation among unknown objects used between training and testing.

%% file: cvpr.bbl
\begin{thebibliography}{10}\itemsep=-1pt

\bibitem{brox2010object}
Thomas Brox and Jitendra Malik.
\newblock Object segmentation by long term analysis of point trajectories.
\newblock In {\em European conference on computer vision}, pages 282--295.
  Springer, 2010.

\bibitem{bucher2019zero}
Maxime Bucher, VU Tuan-Hung, Matthieu Cord, and Patrick P{\'e}rez.
\newblock Zero-shot semantic segmentation.
\newblock In {\em Advances in Neural Information Processing Systems}, pages
  468--479, 2019.

\bibitem{caelles20192019}
Sergi Caelles, Jordi Pont-Tuset, Federico Perazzi, Alberto Montes,
  Kevis-Kokitsi Maninis, and Luc Van~Gool.
\newblock The 2019 davis challenge on vos: Unsupervised multi-object
  segmentation.
\newblock {\em arXiv preprint arXiv:1905.00737}, 2019.

\bibitem{chen2015deepdriving}
Chenyi Chen, Ari Seff, Alain Kornhauser, and Jianxiong Xiao.
\newblock Deepdriving: Learning affordance for direct perception in autonomous
  driving.
\newblock In {\em Proceedings of the IEEE international conference on computer
  vision}, pages 2722--2730, 2015.

\bibitem{cordts2016cityscapes}
Marius Cordts, Mohamed Omran, Sebastian Ramos, Timo Rehfeld, Markus Enzweiler,
  Rodrigo Benenson, Uwe Franke, Stefan Roth, and Bernt Schiele.
\newblock The cityscapes dataset for semantic urban scene understanding.
\newblock {\em arXiv preprint arXiv:1604.01685}, 2016.

\bibitem{dosovitskiy2017carla}
Alexey Dosovitskiy, German Ros, Felipe Codevilla, Antonio Lopez, and Vladlen
  Koltun.
\newblock Carla: An open urban driving simulator.
\newblock {\em arXiv preprint arXiv:1711.03938}, 2017.

\bibitem{geng2020recent}
Chuanxing Geng, Sheng-jun Huang, and Songcan Chen.
\newblock Recent advances in open set recognition: A survey.
\newblock {\em IEEE transactions on pattern analysis and machine intelligence},
  2020.

\bibitem{godard2019digging}
Cl{\'e}ment Godard, Oisin Mac~Aodha, Michael Firman, and Gabriel~J Brostow.
\newblock Digging into self-supervised monocular depth estimation.
\newblock In {\em Proceedings of the IEEE international conference on computer
  vision}, pages 3828--3838, 2019.

\bibitem{he2016deep}
Kaiming He, Xiangyu Zhang, Shaoqing Ren, and Jian Sun.
\newblock Deep residual learning for image recognition.
\newblock In {\em Proceedings of the IEEE conference on computer vision and
  pattern recognition}, pages 770--778, 2016.

\bibitem{ilg2017flownet}
Eddy Ilg, Nikolaus Mayer, Tonmoy Saikia, Margret Keuper, Alexey Dosovitskiy,
  and Thomas Brox.
\newblock Flownet 2.0: Evolution of optical flow estimation with deep networks.
\newblock In {\em Proceedings of the IEEE conference on computer vision and
  pattern recognition}, pages 2462--2470, 2017.

\bibitem{kim2020video}
Dahun Kim, Sanghyun Woo, Joon-Young Lee, and In~So Kweon.
\newblock Video panoptic segmentation.
\newblock In {\em Proceedings of the IEEE/CVF Conference on Computer Vision and
  Pattern Recognition}, pages 9859--9868, 2020.

\bibitem{kirillov2019panoptic}
Alexander Kirillov, Kaiming He, Ross Girshick, Carsten Rother, and Piotr
  Doll{\'a}r.
\newblock Panoptic segmentation.
\newblock In {\em Proceedings of the IEEE conference on computer vision and
  pattern recognition}, pages 9404--9413, 2019.

\bibitem{lin2017feature}
Tsung-Yi Lin, Piotr Doll{\'a}r, Ross Girshick, Kaiming He, Bharath Hariharan,
  and Serge Belongie.
\newblock Feature pyramid networks for object detection.
\newblock In {\em Proceedings of the IEEE conference on computer vision and
  pattern recognition}, pages 2117--2125, 2017.

\bibitem{lin2017focal}
Tsung-Yi Lin, Priya Goyal, Ross Girshick, Kaiming He, and Piotr Doll{\'a}r.
\newblock Focal loss for dense object detection.
\newblock In {\em Proceedings of the IEEE international conference on computer
  vision}, pages 2980--2988, 2017.

\bibitem{milletari2016v}
Fausto Milletari, Nassir Navab, and Seyed-Ahmad Ahmadi.
\newblock V-net: Fully convolutional neural networks for volumetric medical
  image segmentation.
\newblock In {\em 2016 fourth international conference on 3D vision (3DV)},
  pages 565--571. IEEE, 2016.

\bibitem{mohamed2020instancemotseg}
Eslam Mohamed, Mahmoud Ewaisha, Mennatullah Siam, Hazem Rashed, Senthil
  Yogamani, and Ahmad El-Sallab.
\newblock Instancemotseg: Real-time instance motion segmentation for autonomous
  driving.
\newblock {\em arXiv preprint arXiv:2008.07008}, 2020.

\bibitem{ovsep20204d}
Aljo{\v{s}}a O{\v{s}}ep, Paul Voigtlaender, Mark Weber, Jonathon Luiten, and
  Bastian Leibe.
\newblock 4d generic video object proposals.
\newblock In {\em 2020 IEEE International Conference on Robotics and Automation
  (ICRA)}, pages 10031--10037. IEEE, 2020.

\bibitem{Perazzi2016}
F. Perazzi, J. Pont-Tuset, B. McWilliams, L. {Van Gool}, M. Gross, and A.
  Sorkine-Hornung.
\newblock A benchmark dataset and evaluation methodology for video object
  segmentation.
\newblock In {\em Computer Vision and Pattern Recognition}, 2016.

\bibitem{Pont-Tuset_arXiv_2017}
Jordi Pont-Tuset, Federico Perazzi, Sergi Caelles, Pablo Arbel\'aez, Alexander
  Sorkine-Hornung, and Luc {Van Gool}.
\newblock The 2017 davis challenge on video object segmentation.
\newblock {\em arXiv:1704.00675}, 2017.

\bibitem{Rashed_2019_ICCV_Workshops}
Hazem Rashed, Mohamed Ramzy, Victor Vaquero, Ahmad El~Sallab, Ganesh Sistu, and
  Senthil Yogamani.
\newblock Fusemodnet: Real-time camera and lidar based moving object detection
  for robust low-light autonomous driving.
\newblock In {\em The IEEE International Conference on Computer Vision (ICCV)
  Workshops}, Oct 2019.

\bibitem{siam2017modnet}
Mennatullah Siam, Heba Mahgoub, Mohamed Zahran, Senthil Yogamani, Martin
  Jagersand, and Ahmad El-Sallab.
\newblock Modnet: Moving object detection network with motion and appearance
  for autonomous driving.
\newblock {\em arXiv preprint arXiv:1709.04821}, 2017.

\bibitem{snell2017prototypical}
Jake Snell, Kevin Swersky, and Richard Zemel.
\newblock Prototypical networks for few-shot learning.
\newblock In {\em Advances in neural information processing systems}, pages
  4077--4087, 2017.

\bibitem{BMVC.24.56}
David Tsai, Matthew Flagg, and James Rehg.
\newblock Motion coherent tracking with multi-label mrf optimization.
\newblock In {\em Proceedings of the British Machine Vision Conference}, pages
  56.1--56.11. BMVA Press, 2010.
\newblock doi:10.5244/C.24.56.

\bibitem{varma2019idd}
Girish Varma, Anbumani Subramanian, Anoop Namboodiri, Manmohan Chandraker, and
  CV Jawahar.
\newblock Idd: A dataset for exploring problems of autonomous navigation in
  unconstrained environments.
\newblock In {\em 2019 IEEE Winter Conference on Applications of Computer
  Vision (WACV)}, pages 1743--1751. IEEE, 2019.

\bibitem{Valada_2017_IROS}
Johan Vertens, Abhinav Valada, and Wolfram Burgard.
\newblock Smsnet: Semantic motion segmentation using deep convolutional neural
  networks.
\newblock In {\em Proceedings of the IEEE International Conference on
  Intelligent Robots and Systems (IROS)}, Vancouver, Canada, 2017.

\bibitem{Voigtlaender2019CVPR}
Paul Voigtlaender, Michael Krause, Aljosa Osep, Jonathon Luiten, Berin
  Balachandar~Gnana Sekar, Andreas Geiger, and Bastian Leibe.
\newblock Mots: Multi-object tracking and segmentation.
\newblock In {\em Conference on Computer Vision and Pattern Recognition
  (CVPR)}, 2019.

\bibitem{wang2019panet}
Kaixin Wang, Jun~Hao Liew, Yingtian Zou, Daquan Zhou, and Jiashi Feng.
\newblock Panet: Few-shot image semantic segmentation with prototype alignment.
\newblock In {\em Proceedings of the IEEE/CVF International Conference on
  Computer Vision}, pages 9197--9206, 2019.

\bibitem{wang2019solo}
Xinlong Wang, Tao Kong, Chunhua Shen, Yuning Jiang, and Lei Li.
\newblock Solo: Segmenting objects by locations.
\newblock {\em arXiv preprint arXiv:1912.04488}, 2019.

\bibitem{winkens2020contrastive}
Jim Winkens, Rudy Bunel, Abhijit~Guha Roy, Robert Stanforth, Vivek Natarajan,
  Joseph~R Ledsam, Patricia MacWilliams, Pushmeet Kohli, Alan Karthikesalingam,
  Simon Kohl, et~al.
\newblock Contrastive training for improved out-of-distribution detection.
\newblock {\em arXiv preprint arXiv:2007.05566}, 2020.

\bibitem{wong2020identifying}
Kelvin Wong, Shenlong Wang, Mengye Ren, Ming Liang, and Raquel Urtasun.
\newblock Identifying unknown instances for autonomous driving.
\newblock In {\em Conference on Robot Learning}, pages 384--393. PMLR, 2020.

\end{thebibliography}
